\documentclass[letterpaper,10pt]{article}

\usepackage[utf8]{inputenc} 
\usepackage[T1]{fontenc}    
\usepackage{url}            
\usepackage{booktabs}       
\usepackage{amsfonts}       
\usepackage{nicefrac}       
\usepackage{microtype}      
\usepackage{xcolor}         
\usepackage{booktabs} 
\usepackage{setspace}
\usepackage{amsmath}
\usepackage{subfig}
\usepackage{multirow}
\usepackage{tikz}
\usetikzlibrary{bayesnet}
\usepackage{graphicx}
\usepackage{wrapfig}
\usepackage{float}
\usepackage{booktabs} 
\usepackage{lipsum}   
\usepackage{tikz}
\usepackage{authblk}
\usepackage{fancyhdr}
\usepackage{xcolor}
\usepackage{url}
\usepackage{enumitem}
\usepackage{amssymb} 
\usepackage{pifont}

\usepackage{probml}

\ShortHeadings{Hierarchical Bayesian Crowdsourcing}

\usepackage{blindtext}
\usepackage{wrapfig}

\usepackage{bm,amssymb,amsmath}

\makeatletter
\def\th@plain{%
  \thm@notefont{}
  \itshape 
}
\def\th@definition{%
  \thm@notefont{}
  \normalfont 
}
\makeatother

\def\1{\bm{1}}

\DeclareMathAlphabet{\mathsfit}{\encodingdefault}{\sfdefault}{m}{sl}
\SetMathAlphabet{\mathsfit}{bold}{\encodingdefault}{\sfdefault}{bx}{n}

\begin{document}

\title{
   Hierarchical Bayesian Crowdsourcing \\ with Item Difficulty
}

\author[1,$\dagger$]{Seong Woo Han}
\author[2]{Ozan Adıgüzel}
\author[3]{Bob Carpenter}

\affil[1]{University of Pennsylvania}
\affil[2]{Columbia University}
\affil[3]{Center for Computational Mathematics, Flatiron Institute}
\affil[$\dagger$]{Correspondence to \url{seonghan@seas.upenn.edu}}

\maketitle

\begin{abstract}
In applied statistics and machine learning, the gold standards used for training are often biased and almost always noisy. Dawid and Skene's justifiably popular crowdsourcing model adjusts for rater sensitivity and specificity, but fails to capture distributional properties of rating data gathered for training, which in turn biases training. In this study, we introduce a general purpose measurement-error model with which we can infer consensus categories by adding item-level effects for difficulty, discriminativeness, and guessability. We further show how to constrain the bimodal posterior of these models to avoid adversarial raters. We validate our model's goodness of fit with posterior predictive checks, the Bayesian analogue of $\chi^2$ tests, and assess its predictive accuracy using leave-one-out cross-validation. We illustrate our new model with two well-studied data sets, binary rating data for caries in dental X-rays and implication in natural language. 
\end{abstract}

\section{Introduction}
\begingroup
\let\thefootnote\relax\footnotetext{We provide code at \url{https://github.com/seongwoohan/bayes-crowdsourcing-irt}.}
\endgroup

Crowdsourcing is the process of soliciting ratings (labels, annotations, codes) from a group of respondents for a set of items and combining them for downstream tasks such as training a neural network classifier or a large language model. This note focuses on binary ratings arising in two-way classification problems. In practice, such ratings are often noisy, biased, and inconsistent across raters and items. For example, dentists labeling dental X-rays for caries show low agreement and systematic differences in labeling tendencies \citep{espeland1989}, and annotators in natural language inference datasets exhibit substantial variability \citep{snow2008}. As a result, heuristic approaches such as majority voting ignore uncertainty and can lead to biased estimates and degraded performance. Several probabilistic models have been proposed to describe the annotation process. The simplest multinomial model \citet{albert2004} assumes equal annotator reliability and leads to equal-weight voting. \citet{dawid1979} introduced annotator-specific sensitivity and specificity, providing a widely used framework for modeling rater reliability. Subsequent work incorporates limited item structure; for example, \citet{beigman2009learning} model items as easy or regular, with annotators agreeing on easy items. However, these models do not capture continuous variation in item properties such as difficulty, discrimination, and guessing, nor do they address identifiability issues arising from adversarial raters. More recent applications, such as human feedback for fine-tuning large language models \citet{ouyang2022, rafailov2024}, further highlight the need for flexible and interpretable models of noisy annotation; additional applications are discussed in Appendix~\ref{apd:app}.

In this work we ask how crowdsourcing models can be extended to better capture the heterogeneity of raters and items while remaining identifiable. To answer this question, the main contributions of the paper are: (i) we propose a hierarchical Bayesian measurement-error model for crowdsourced binary ratings with item-level difficulty, discrimination, and guessing effects, (ii) we resolve the bimodality of rating models by introducing a constraint that excludes adversarial raters, improving identifiability while retaining flexibility near the spammy boundary, and (iii) we provide open-source implementations in Stan \citep{carpenter2017}, enabling use in R, Python, and Julia.

\section{Hierarchical Bayesian crowdsourcing model}

\paragraph{Generating ratings.}
We consider $I$ items and $J$ raters with $N$ binary ratings $y_n \in \{0,1\}$ in long format, where each rating is associated with an item $\text{item}_n \in \{1,\ldots,I\}$ and rater $\text{rater}_n \in \{1,\ldots,J\}$. Each item $i$ has a latent category $z_i \in \{0,1\}$ with prevalence $\pi \in (0,1)$, $z_i \sim \mathrm{Bernoulli}(\pi)$. For estimation, we marginalize over $z_i$ to form the likelihood used in inference; details are in Appendix~\ref{apd:first}.

Ratings are generated conditionally on $z_i$. Each rater $j$ is characterized by sensitivity and specificity parameters $\alpha^{\textrm{sens}}_j, \alpha^{\textrm{spec}}_j \in \mathbb{R}$ on the log-odds scale, governing accuracy on positive and negative items, respectively. Each item $i$ is parameterized by difficulty $\beta_i \in \mathbb{R}$, discrimination $\delta_i > 0$, and guessability $\lambda_i \in (0,1)$. Difficulty shifts the log-odds of a correct response, so higher $\beta_i$ corresponds to harder items; discrimination scales this difference, so higher $\delta_i$ makes responses more sensitive to the gap between rater ability and item difficulty; and guessability $\lambda_i$ captures the probability of a correct response due to guessing, ensuring a nonzero baseline accuracy even for very difficult items. The full, unreduced model follows the item-response theory three-parameter logistic model generalized with sensitivity and specificity (which we denote IRT-3PL, despite the generalization), where the probability that rater $j$ assigns the correct rating to item $i$ is given by
\begin{equation}
c_n \sim \textrm{bernoulli}\!\left(\lambda_i + (1 - \lambda_i)\,\textrm{logit}^{-1}\!\left(\delta_i(\alpha^k_j - \beta_i)\right)\right),
\end{equation}
where $k = \textrm{sens}$ if $z_i = 1$ and $k = \textrm{spec}$ if $z_i = 0$. To obtain a distribution over observed ratings, the probability is flipped for negative items so that high accuracy corresponds to a high probability of the correct label. Thus, the rating is given by
\begin{equation}
\begin{split}
y_n \sim
\begin{cases}
    \textrm{Bernoulli} \left( \lambda_i + (1 - \lambda_i)  \right. \textrm{logit}^{-1} \left( \delta_i (\alpha_j^{\textrm{sens}} - \beta_i) \right) \left. \right), & \text{if } z_i = 1, \\
 \\[-8pt]
    \textrm{Bernoulli} \left( 1 - \left( \lambda_i + (1 - \lambda_i) \right. \right. \textrm{logit}^{-1} \left( \delta_i (\alpha_j^{\textrm{spec}} - \beta_i) \right) \left. \right) \left. \right), & \text{if } z_i = 0.
\end{cases}
\end{split}
\end{equation}
\normalsize

The second case ($z_i = 0$) reduces to the expression $\textrm{bernoulli}\!\left( (1 - \lambda_i)\cdot \left( 1 - \textrm{logit}^{-1}\!\left(\delta_i \cdot \left(\alpha^\textrm{spec}_j - \beta_i\right)\right)\right)\right).$

\paragraph{Rater model and priors.}
A spammy rater is one for whom the rating does not depend on the item being rated. This occurs when $\alpha^{\textrm{sens}}_j = -\alpha^{\textrm{spec}}_j$, so that on the probability scale $\textrm{logit}^{-1}(\alpha^{\textrm{sens}}_j) = 1 - \textrm{logit}^{-1}(\alpha^{\textrm{spec}}_j)$ \citet{passonneau2014}. For example, a rater with 40\% sensitivity and 60\% specificity has a 40\% chance of returning a positive rating regardless of the true category.

An adversarial rater is one for which $\alpha^{\textrm{sens}}_j < -\alpha^{\textrm{spec}}_j$, implying sensitivity is less than one minus specificity. Such raters are systematically more likely to be incorrect than correct. For example, raters with sensitivity and specificity below $50\%$ are adversarial, whereas those above $50\%$ are informative. To ensure identifiability and avoid symmetric likelihood contributions, we exclude adversarial behavior by reparameterizing sensitivity while retaining flexibility near the spammy boundary, 
\begin{equation}
\alpha^{\textrm{sens}}_j = -\alpha^{\textrm{spec}}_j + \log(1+\exp(\eta_j)), 
\end{equation}
where $\eta_j \in \mathbb{R}$. This ensures $\alpha^{\textrm{sens}}_j > -\alpha^{\textrm{spec}}_j$ for all raters, while allowing raters arbitrarily close to the spammy regime. We complete the model by placing weakly informative priors on all parameters to regularize the scale of the parameters and aid identifiability, as recommended by \citet{gabry2019visualization} and \citet{gelman2020bayesian}. For item-level parameters, $\pi \sim \text{Beta}(2,2)$, $\beta_i \sim \text{Normal}(0,1)$, $\delta_i \sim \text{LogNormal}(0,0.25)$, and $\lambda_i \sim \text{Beta}(2,2)$. To model heterogeneity across raters, we place a hierarchical bivariate normal prior on $(\eta_j, \alpha^{\textrm{spec}}_j)$, where $\eta_j$ controls the distance from the spammy boundary, 
\begin{equation}
\begin{pmatrix} \eta_j \\ \alpha^{\textrm{spec}}_j \end{pmatrix} \sim \mathrm{MVN}\!\left(\mu_\alpha,\ \mathrm{diag}(\tau_\alpha)\,\Omega_\alpha\,\mathrm{diag}(\tau_\alpha)\right), 
\end{equation}
with $\mu_\alpha \sim \mathrm{Normal}(0,5)$, $\tau_\alpha \sim \mathrm{LogNormal}(0,1)$, and $\Omega_\alpha \sim \mathrm{LKJ}(2)$. For posterior computation, we use the equivalent non-centered reparameterization \citet{papaspiliopoulos2007general} to decouple latent variables from their scales and mitigates Neal’s funnel \citet{neal2003slice} for faster and more stable convergence: 
\begin{equation}
    \begin{pmatrix} \eta_j \\ \alpha^{\textrm{spec}}_j \end{pmatrix}
= \mu_\alpha + \mathrm{diag}(\tau_\alpha)L_{\Omega_\alpha} z_j,\quad
z_j \sim \mathcal{N}(\mathbf{0}, I_2).
\end{equation}

\begin{wraptable}{r}{0.5\textwidth}
\begin{tabular}{c c l}
\toprule
Tag & Reduction & Description \\ 
\midrule
A & \( \lambda_i = 0 \) & no guessing items \\ 
B & \( \delta_i = 1 \) & equal discrimination items \\ 
C & \( \beta_i = 0 \) & equal difficulty items \\ 
D & \( \alpha^{\text{spec}} = \alpha^{\text{sens}} \) & equal error raters \\ 
E & \( \alpha_i = \alpha_j \) & identical raters \\ 
\bottomrule
\end{tabular}
\caption{Orthogonal model reductions.}
\label{table:models1}
\end{wraptable}

\paragraph{Model reductions.}
By tying or fixing parameters, the full (IRT 3PL $+$ sens/spec) model induces a range of natural submodels, including standard IRT variants and the Dawid–Skene model. Table~\ref{table:models1} summarizes the reduction structure via parameter constraints. We consider selected reductions of the full model. Models D, BD, and ABD correspond to tied sensitivity and specificity (i.e., $\alpha^{\text{spec}} = \alpha^{\text{sens}}$), while the remaining models allow separate sensitivity and specificity parameters. Additional model reductions are provided in Appendix~\ref{apd:second}.

\begin{equation}
\setlength{\arraycolsep}{2pt}
\begin{array}{ccl}
\textrm{Reductions} & \textrm{Probability Correct} & \textrm{Note} \\ \hline
 & \lambda_i + (1 - \lambda_i)\,\textrm{logit}^{-1}(\delta_i (\alpha^k_j - \beta_i)) & \textrm{\small IRT 3PL + sens/spec}
\\
A
& \textrm{logit}^{-1}(\delta_i (\alpha^k_j - \beta_i)) & \textrm{\small IRT 2PL + sens/spec}
\\
AB
& \textrm{logit}^{-1}(\alpha^k_j - \beta_i) & \textrm{\small IRT 1PL + sens/spec}
\\
ABC
& \textrm{logit}^{-1}(\alpha^k_j) & \textrm{Dawid--Skene}
\\ \hline
D
& \lambda_i + (1 - \lambda_i)\,\textrm{logit}^{-1}(\delta_i (\alpha_j - \beta_i)) & \textrm{\small IRT 3PL}
\\
BD
& \lambda_i + (1 - \lambda_i)\,\textrm{logit}^{-1}(\alpha_j - \beta_i) & \textrm{\small IRT 2PL}
\\
ABD
& \textrm{logit}^{-1}(\alpha_j - \beta_i) & \textrm{\small IRT 1PL}
\end{array}
\end{equation}

\begin{figure}[t]
  \centering
  \includegraphics[width=0.75\textwidth]{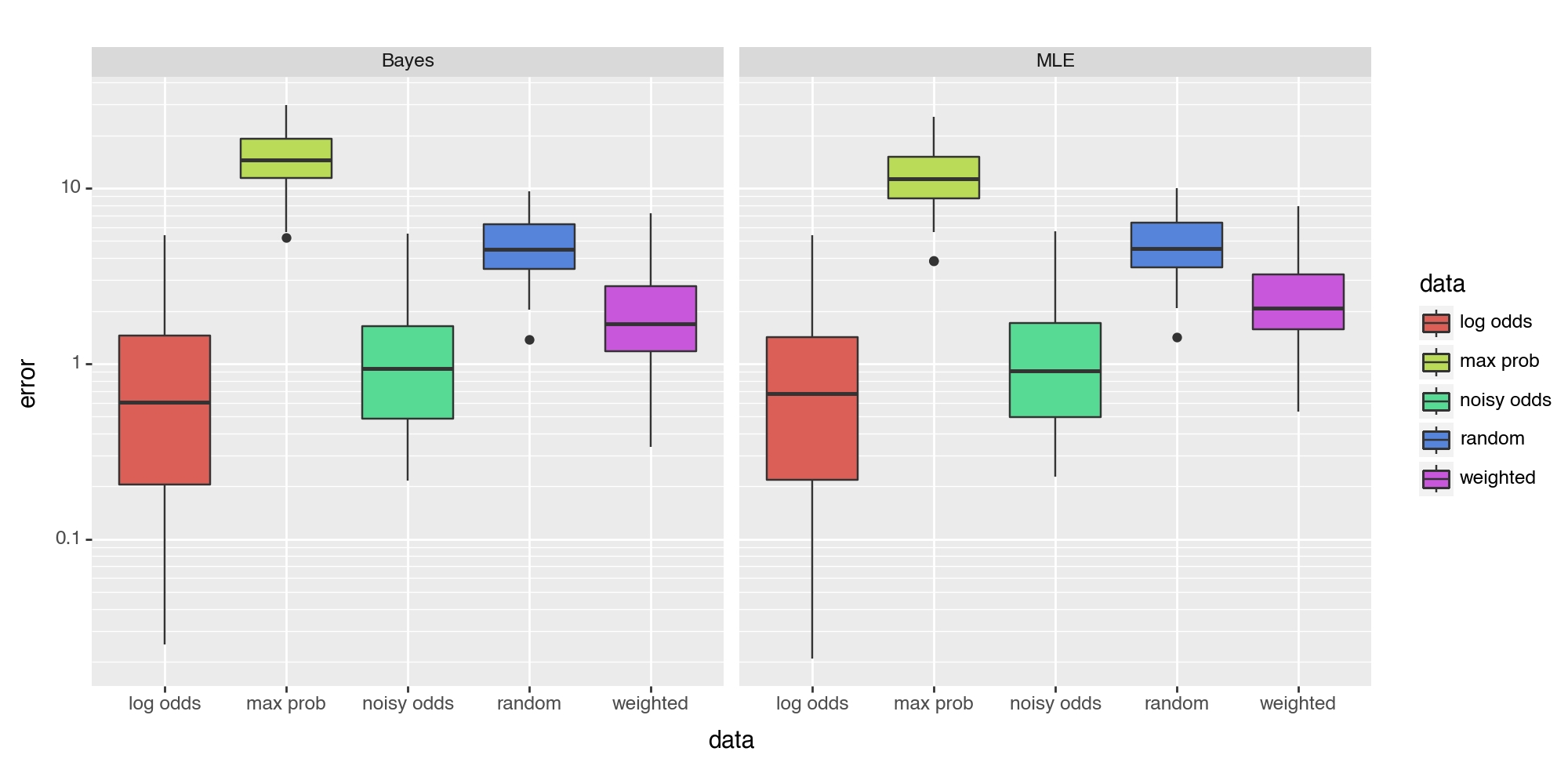}
  \caption{The L2 norm of parameter estimation error, $||\widehat{\theta} - \theta||_2$, for different approaches to training with probabilities (lower is better): (log odds) training a linear regression on the log odds, (max prob) assigning the highest probability category, (noisy odds) add standard normal noise to the log odds approach, (random) assign a random category according to the probability, (weighted) train a weighted logistic regression. Estimates and variability are consistent between Bayesian posterior means with a normal prior (left) or ridge-penalized maximum likelihood estimates (right).  Regression is 32-dimensional, with correlated inputs and 1024 training data points.  Results show standard bar-and-whisker plots over 32 trials with paired random $x, \beta$.}\label{fig:prob-train-err}
\end{figure}

\section{Training on probabilistic data}

Rating models produce posterior probabilities $\Pr[y_n = 1]$ rather than hard labels. Using these probabilities directly for training preserves uncertainty and avoids bias introduced by collapsing to binary gold standard labels. Consider a classification setting with covariates $x_n \in \mathbb{R}^L$ and binary outcomes $y_n \in \{0,1\}$, modeled via logistic regression, $y_n \sim \textrm{bernoulli}\!\left(\textrm{logit}^{-1}(x_n \cdot \beta)\right)$. In practice, $y_n$ is unobserved and replaced by an estimate of $\Pr[y_n = 1]$, whereas standard approaches instead use hard labels obtained by thresholding or majority vote. 

To evaluate the impact of probabilistic training, we conduct a synthetic experiment (details in Appendix~\ref{apd:third}). The resulting error norms in estimating $\beta$ are shown in Figure~\ref{fig:prob-train-err}. The plot shows that training with the probabilities is much better than taking the category with the highest probability. The best approach is training a linear regression based on the log odds, with the noisy version of the same approach not far behind. Weighted training with logistic regression is not quite as good.  Randomly selecting a category according to the generative model is not as good as using the weights directly, but it still dominates taking the best category.

\section{Model evaluation}

\begin{figure}[t]
  \centering
  \includegraphics[width=1.0\linewidth]{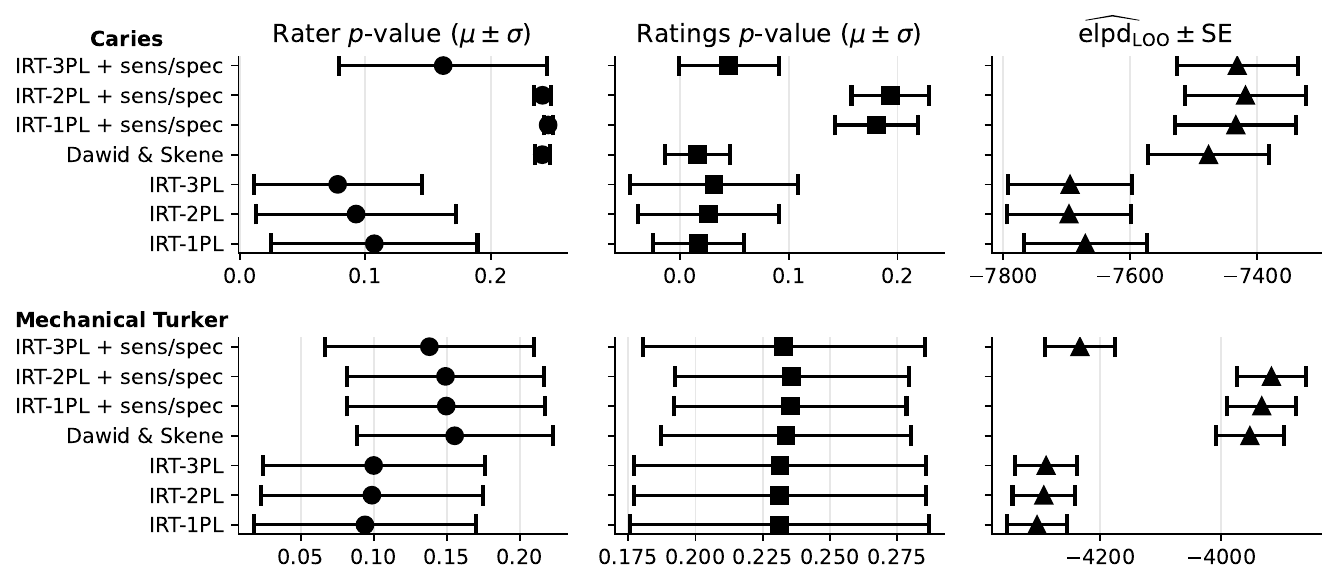}
  \caption{Posterior predictive $p$-values and leave-one-out cross-validation $\text{elpd}_{\text{loo}}$ across Caries and Mechanical Turk datasets for 7 models. Posterior predictive values are summarized by mean and standard deviation, while $\text{elpd}_{\text{loo}}$ values are reported with standard errors.}
  \label{fig:ppc_loo}
\end{figure}

The posteriors of all 7 distinct models were estimated using the No-U-Turn sampler, an adaptive Hamiltonian Monte Carlo method \citep{hoffman2014, betancourt2017}, and evaluated on two datasets using posterior predictive checks (PPC) for goodness-of-fit and leave-one-out cross-validation (LOO) for predictive accuracy \citep{gelman1996, vehtari2017practical} (details in Appendix~\ref{apd:fourth}). The first consists of 5 dentists rating roughly 4000 dental X-rays for caries \citep{espeland1989}, and the second involves nearly 200 Mechanical Turkers rating subsets of roughly 3000 sentence pairs for entailment \citep{snow2008}. Models were implemented in Stan (v2.33) and fit using CmdStanPy (v1.20) with default settings. All runs used four chains with 1000 warmup and 1000 sampling iterations, and achieved split-$\widehat{R} < 1.01$ for all parameters, indicating convergence \citep{gelman2013}. Figure~\ref{fig:ppc_loo} shows posterior predictive $p$-values and $\text{elpd}_{\text{loo}}$ for all models, where values are summarized by their mean and standard deviation for posterior predictive checks and by standard errors for leave-one-out cross-validation, with error bars reflecting simulation variability and estimation uncertainty, respectively. Posterior predictive $p$-values assess how well model-simulated ratings match the observed data, with values near 0.5 indicating good fit. Among models that pass these checks, elpd\textsubscript{loo} provides a measure of predictive accuracy, where higher values indicate better performance. Results for the full set of 18 model reductions (Appendix~\ref{apd:second}) are provided in Appendix~\ref{apd:fifth}.

For the caries data, the IRT-2PL + sens/spec provides the best overall performance, achieving the highest $\text{elpd}_{\text{loo}}$. The IRT-1PL + sens/spec and IRT-3PL + sens/spec models perform similarly but slightly worse in predictive accuracy. Posterior predictive checks reinforce this pattern. While all sens/spec models provide reasonable fits, the IRT-3PL + sens/spec model exhibits increased variability, suggesting that the additional guessing parameter introduces  complexity and is not well supported by the data. In contrast, models without sensitivity and specificity (i.e., standard IRT 1PL--3PL) perform substantially worse in both posterior predictive fit and predictive accuracy. The Dawid--Skene model, in particular, fails to capture the ratings distribution adequately, with poor posterior predictive behavior for rating-level summaries. Overall, these results demonstrate that explicitly modeling rater sensitivity and specificity is critical, providing an improvement over both classical IRT models and Dawid--Skene. For the Mechanical Turk data, the IRT-2PL + sens/spec again achieves the best predictive performance in terms of $\text{elpd}_{\text{loo}}$. Posterior predictive checks show that the rater $p$-values are broadly similar across the IRT models with sensitivity and specificity and the Dawid--Skene model. However, models incorporating sensitivity and specificity consistently outperform those without it in predictive accuracy. None of the models provide a strong fit for the ratings $p$-values. The main difference between the Caries and Mechanical Turk datasets is that there are many more Turkers providing far fewer ratings each, and there are more spammy or low-quality raters among the Turkers, which makes the inference problem substantially noisier.

\section{Dawid and Skene’s posterior predictions underestimate dispersion}

\begin{wrapfigure}{r}
{0.57\textwidth}
  \centering
  \includegraphics[width=0.435\textwidth]{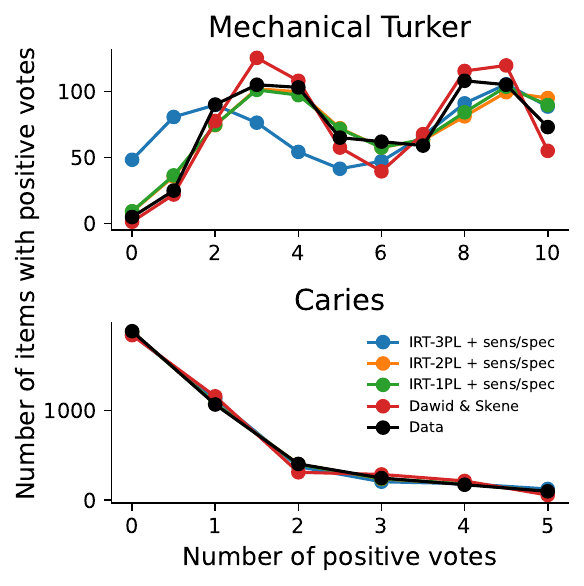}
  \caption{Distribution of positive votes per item, demonstrating the varying levels of dispersion captured by each model.}
  \label{fig:understimate}
\end{wrapfigure}

To demonstrate how posterior predictions underestimate dispersion, we compare Dawid and Skene's baseline with three item-level models: difficulty, plus discrimination, and plus guessing with sensitivity and specificity in Amazon Turker data. The goodness-of-fit tests are similar, so in Figure~\ref{fig:understimate}, we plot the expected number of marginal positive votes for each model along with the actual data. The Dawid and Skene model has an inflated number of middling positive votes compared to the data; in contrast the IRT-3PL + sens/spec model underestimates them.  However, posterior predictive inference for IRT-2PL + sens/spec and IRT-1PL + sens/spec more closely matches the actual data. This alignment enhances the model's ability to reflect true voter consensus, thereby producing a distribution of votes that more accurately mirrors observed voting patterns. For the caries data, the story is similar. IRT-1,2,3PL + sens/spec closely match the actual data, while the Dawid and Skene model again inflates the number of middling positive votes compared to the real data.


\section{Conclusion}
Until this work, crowdsourcing models for categorical responses, such as Dawid--Skene \citep{dawid1979}, do not account for item-level variation or structured rater behavior, leading to systematic misfit. Our results demonstrate this through posterior predictive checks and leave-one-out cross-validation, where Dawid--Skene fails to capture observed rating patterns. By incorporating item difficulty, discrimination, and guessability within a hierarchical Bayesian framework, our models achieve improved fit and predictive performance. Explicit modeling of rater sensitivity and specificity further yields more accurate and well-calibrated predictions. To ensure identifiability, we restrict to cooperative solutions, excluding adversarial modes where raters perform below chance, which are less realistic in typical crowdsourcing settings \citep{passonneau2014}. Potential extensions of the current framework are discussed in Appendix~\ref{apd:extension}. The bottom line is that most training efforts in machine learning are not making the most of their human feedback and can be improved by applying the structured probabilistic crowdsourcing models.




\clearpage



\bibliography{references}

\clearpage

\appendix

\section{Applications of crowdsourcing models}\label{apd:app}
There are several tasks to which crowdsourcing models are applied, and for every one of them a rating model improves performance over heuristic baselines.  For example, rating models outperform majority voting for category training and outperform indirect measurements like inter-annotator agreement statistics to measure task difficulty (see, e.g., \cite{artstein2008inter}, \cite{sabou2014corpus}, \cite{mchugh2012interrater}).

\textbf{Inferring a gold-standard data set.} The first and foremost application of crowdsourcing is to generate a ``gold standard'' data set, where a single category (or label) is assigned to each item.  In terms of generating representative data, it is best to sample data according to its probability (i.e., follow the generative model) rather than to choose the ``best'' rating for each item according to a heuristic such as highest probability.  The second section of this paper shows that it is better for downstream accuracy to train with a probabilistic corpus that retains information about rating uncertainty.  In cases where it is impractical to use a probabilistic corpus with weighted training, we show why it is far better to sample labels according to their posterior probability in the rating model than to choose the ``best'' label.  In particular, we will demonstrate that majority voting schemes among raters are suboptimal compared to sampling, which is in turn dominated by training with the probabilities (a kind of Rao-Blackwellization).  If a classifier for the data is available, an even better approach is to jointly train a classifier and rating model, as shown by \cite{raykar2010}.

\textbf{Inferring population prevalence.} The second most common application of crowdsourcing is to understand the probability of positivity among items in the population represented by the crowdsourcing data.  This is particularly common in epidemiology, where the probability of positive outcomes is the prevalence of the disease in the (sub)population \citep{albert2004}.  It can also be used to analyze the prevalence of hate speech on a social media site or bias in televised news, the prevalence of volcanoes on Venus \citep{smyth1994}, or the prevalence of positive reviews for a restaurant.

\textbf{Understanding and improving the coding standard.} The third most common application of crowdsourcing is to understand the coding standard, which is the rubric under which rating is carried out.  Traditionally, this has been measured through inter-annotator agreement \citep{artstein2008inter}.  In contrast, rating models provide finer-grained analysis of rater accuracy in terms of sensitivity and specificity, as well as the information gain expected from a rating by a specific rater \citep{passonneau2014}.  

\textbf{Understanding and improving raters.} What is the mean sensitivity and specificity and how does it vary among raters?  Are sensitivity and specificity anticorrelated or correlated in the population? This understanding can be fed back to the raters themselves for ongoing training.  For example, American baseball umpires have extensive feedback on how they call balls and strikes as measured against a very accurate machine's call, which has led to much higher accuracy and consistency among umpires \cite{flannagan2024}.  Understanding rater populations, such as those available through Mechanical Turk or Upwork, is important when managing raters for multiple crowdsourcing tasks.  For example, it is straightforward with a rating model to infer the proportion of spammers and the proportion of high quality raters.  The number of raters required for high quality joint ratings may also be assessed \citep{passonneau2014}.  

\textbf{Understanding the items.} A fifth task, which has received relatively little attention in the crowdsourcing literature, is to understand the structure of the population of items. For example, which items are difficult to rate and why? Which items simply have too little signal to be consistently rated?  Which items lie near the decision boundary and which are far away? Which items have high discrimination and why? A discriminative item is one which cleanly separates high ability from low ability raters in their ability to rate it correctly.  Understanding the items is the primary focus of educational test design, where the items are test questions and the raters are students.  Test questions are selected for standardized tests like the ACT (American College Testing) based on having high discrimination and a useful range of difficulties \citep{act2020}.

\section{Marginal likelihood}\label{apd:first}
A crowdsourcing model generates latent discrete categories $z_i \in \{ 0, 1 \}$ for each item.  For both optimization and sampling, it is convenient to marginalize the complete likelihood $p(y, z \mid \pi, \alpha, \beta, \delta, \lambda)$ to the rating likelihood $p(y \mid \pi, \alpha, \beta, \delta, \lambda)$.  The marginalization calculation is efficient because it is factored by data item.  Letting $\theta = \pi, \alpha, \beta, \delta, \lambda$ be the full set of continuous parameters, the trick is to rearrange the long-form data by item, then marginalize out the discrete parameters, resulting in the likelihood
\begin{equation}
\setlength{\abovedisplayskip}{3pt}
\setlength{\belowdisplayskip}{3pt}
\label{eq:likelihood}
p(y \mid \theta)
= \prod_{i=1}^{I} \sum_{z_i = 0}^1
p(z_i \mid \theta) \cdot\hspace{-13pt} \prod_{n : \textrm{item}_n = i} \hspace{-9pt} p(y_n \mid z_i, \theta).
\end{equation}

We start with the complete data likelihood for ratings $y$ and latent categories $z$, 
\begin{equation}
p(y, z \mid \theta)
=
\prod_{i=1}^I p(z_i \mid \theta)
\cdot
\prod_{n = 1}^N p(y_n \mid z, \theta)
\end{equation}
and then rearrange terms by item,
\begin{equation}
p(y, z \mid \theta) = 
\prod_{i=i}^I
  \left( p(z_i \mid \theta)
         \cdot
	 \prod_{n : \textrm{item}_n = i} p(y_n \mid z_i, \theta)
  \right).
\end{equation}

On a per item basis, the marginalization is tractable, yielding Equation~\ref{eq:likelihood}. Computational inference requires working on the log scale, where the log marginal likelihood of the rating data in the full model is given by
\begin{equation}
\begin{array}{rcl}
\log p(y \mid \theta)
& = & \log \prod_{i=i}^I
  \sum_{z_i = 0}^1 \,
          p(z_i \mid \theta)
                \cdot
		\prod_{n : \textrm{item}_n = i} p(y_n \mid z_i, \theta).
\\[4pt]
& = &
\sum_{i=1}^I
\textrm{logSumExp}_{z_i = 0}^1 \,
  \left(
    \log p(z_i \mid \theta)
    + 
    \sum_{n : \textrm{item}_n = i} \log p(y_n \mid z_i, \theta)
  \right),
\end{array}
\end{equation}
where
\begin{equation}
\textrm{logSumExp}_{n = 1}^N \, \ell_n
= \log \sum_{n=1}^N \exp(\ell_n)
\end{equation}
is the numerically stable log-scale analogue of addition.

\section{Full model reductions}\label{apd:second}

\paragraph{Tied sensitivity and specificity.} First, we consider models which do not distinguish sensitivity and specificity. Such models should be used when the categories are not intrinsically ordered (e.g., rating two consumer brands for preference). All of these models other than the last (ABCDE) assumes raters have varying accuracy. 

\vspace*{-15pt}

\begin{equation}
\setlength{\arraycolsep}{2pt}
\begin{array}{ccl}
\textrm{Reductions} & \textrm{Probability Correct} & \textrm{Note} \\ \hline
D
& \lambda_i + (1 - \lambda_i) \cdot \textrm{logit}^{-1}(\delta_i \cdot (\alpha_j - \beta_i)) & \textrm{\small IRT 3PL}
\\ 
CD
& \lambda_i + (1 - \lambda_i) \cdot \textrm{logit}^{-1}(\delta_i \cdot \alpha_j)
\\
BD
& \lambda_i + (1 - \lambda_i) \cdot \textrm{logit}^{-1}(\alpha_j - \beta_i) & \textrm{\small IRT 2PL}
\\ \hline
BCD
& \lambda_i + (1 - \lambda_i) \cdot \textrm{logit}^{-1}(\alpha_j)
\\
AD
& \textrm{logit}^{-1}(\delta_i \cdot (\alpha_j - \beta_i))
\\
ACD
& \textrm{logit}^{-1}(\delta_i \cdot \alpha_j)
\\ \hline
ABD
& \textrm{logit}^{-1}(\alpha_j - \beta_i) & \textrm{\small IRT 1PL}
\\
ABCD
& \textrm{logit}^{-1}(\alpha_j)
\\
ABCDE
& \textrm{logit}^{-1}(\alpha)
\end{array}
\end{equation}

\paragraph{Free sensitivity and specificity.}  The following models introduce separate parameters for sensitivity and specificity rather than assuming they are the same.  
Only the last model (ABCE) does not distinguish rater abilities.

\vspace*{-15pt}

\begin{equation}
\setlength{\arraycolsep}{2pt} 
\begin{array}{@{}l@{\hskip -10pt}c@{\hskip 4pt}l@{}}
\textrm{Reductions} & \textrm{Probability Correct} & \textrm{Note} \\ \hline
 & \lambda_i + (1 - \lambda_i) \cdot \textrm{logit}^{-1}(\delta_i \cdot (\alpha^k_j - \beta_i)) & \textrm{\small IRT 3PL + sens/spec} \\
C & \lambda_i + (1 - \lambda_i) \cdot \textrm{logit}^{-1}(\delta_i \cdot \alpha^k_j) & \\
BC & \lambda_i + (1 - \lambda_i) \cdot \textrm{logit}^{-1}(\alpha^k_j) & \\
\hline
A & \textrm{logit}^{-1}(\delta_i \cdot (\alpha^k_j - \beta_i)) & \textrm{\small IRT 2PL + sens/spec} \\
AC & \textrm{logit}^{-1}(\delta_i \cdot \alpha^k_j) & \\
AB & \textrm{logit}^{-1}(\alpha^k_j - \beta_i) & \textrm{\small IRT 1PL + sens/spec} \\
\hline
ABC & \textrm{logit}^{-1}(\alpha^k_j) & \textrm{Dawid/Skene} \\ 
ABCE & \textrm{logit}^{-1}(\alpha^k) & \\
\end{array}
\end{equation}

\paragraph{No rater effects.} The last model, which is common in epidemiology \citep{albert2004}, includes item effects without any rater effects.
\begin{equation}
\begin{array}{cc}
\textrm{Reductions} & \textrm{Probability Correct} \\ \hline
ABDE
& \textrm{logit}^{-1}(- \beta_i) \quad
\\ \hline
\end{array}
\end{equation}

\paragraph{Redundant parameter models.} The remaining thirteen models are redundant in the sense that fixing their non-identifiability issues reduces to a model with a single item effect. 

\begin{equation}
\begin{array}{ccc}
\textrm{Reductions} & \textrm{Probability Correct} \\ \hline
E
& \lambda_i + (1 - \lambda_i) \cdot \textrm{logit}^{-1}(\delta_i \cdot (\alpha^k - \beta_i))
\\
DE
& \lambda_i + (1 - \lambda_i) \cdot \textrm{logit}^{-1}(\delta_i \cdot (\alpha - \beta_i))
\\
CE
& \lambda_i + (1 - \lambda_i) \cdot \textrm{logit}^{-1}(\delta_i \cdot \alpha^k)
\\ \hline
CDE
& \lambda_i + (1 - \lambda_i) \cdot \textrm{logit}^{-1}(\delta_i \cdot \alpha)
\\
BE
& \lambda_i + (1 - \lambda_i) \cdot \textrm{logit}^{-1}(\alpha^k - \beta_i)
\\
BDE
& \lambda_i + (1 - \lambda_i) \cdot \textrm{logit}^{-1}(\alpha - \beta_i)
\\ \hline
BCE
& \lambda_i + (1 - \lambda_i) \cdot \textrm{logit}^{-1}(\alpha^k)
\\
BCDE
& \lambda_i + (1 - \lambda_i) \cdot \textrm{logit}^{-1}(\alpha)
\\
AE 
& \textrm{logit}^{-1}(\delta_i \cdot (\alpha^k - \beta_i))
\\ \hline
ADE
& \textrm{logit}^{-1}(\delta_i \cdot (\alpha - \beta_i))
\\
ACE
& \textrm{logit}^{-1}(\delta_i \cdot \alpha^k)
\\
ACDE
& \textrm{logit}^{-1}(\delta_i \cdot \alpha)
\\ \hline
ABE
& \textrm{logit}^{-1}(\alpha^k - \beta_i)
\end{array}
\end{equation}

\section{Synthetic experiment details}\label{apd:third}
For the experiment, we generate a synthetic data set of covariates $x_i \sim \textrm{normal}(0, \Sigma)$, where the positive definite covariance matrix is defined by $\Sigma_{m, n} = \rho^{|m - n|}$, with $\rho = 0.9$.  The result is highly correlated covariates; uncorrelated covariates show the same trend.  We evaluate five approaches to estimation: (max prob) take $y_n$ to be 1 if $\textrm{logit}^{-1}(x_n \cdot \beta) > \frac{1}{2},$ (log odds) train a linear regression with outcome $\log x_n \cdot \beta,$ (noisy odds) log odds with standard normal noise, $\textrm{logit}^{-1}(x_n \cdot \beta + \epsilon_n)$, with $\epsilon_n \sim \textrm{normal}(0, 1),$ (random) randomly generate $y_n \sim \textrm{bernoulli}\!\left(\textrm{logit}^{-1}(x_n \cdot \beta)\right)$ according to its probability distribution, and (weighted) train a weighted logistic regression with outcome 1 and weight $\textrm{logit}^{-1}(x_n \cdot \beta)$ \textit{and}\, outcome 0 and weight $1 - \textrm{logit}^{-1}(x_n \cdot \beta).$

We provide both Bayesian (posterior mean) and frequentist (penalized maximum likelihood) estimates.  For the Bayesian setting, we use a standard normal prior $\beta_k \sim \textrm{normal}(0, 1)$ and in the frequentist setting we use ridge regression, with penalty function $\frac{1}{2} \cdot \beta^\top \cdot \beta$ to match the Bayesian prior.

\section{Evaluation metrics}\label{apd:fourth}
\paragraph{Posterior predictive checks} We use posterior predictive checks (PPC) for goodness-of-fit testing \citep{gelman1996}. They are the Bayesian analogue of the $\chi^2$ tests widely employed in epidemiology \citep{albert2004}, which test a “null” of the fitted model against the observed data \citep{formann2003}. PPCs work by generating replicated data from the fitted model and comparing statistics from the replicated data to those from the original data. The posterior predictive distribution for replications $y^{\textrm{rep}}$ of the original data $y$ given model parameters $\theta$ is $p(y^{\textrm{rep}}\mid y) 
= \mathbb{E}\left[ p\!\left(y^\text{rep} \mid \theta\right) \mid y\right]
= \int p\!\left(y^{\textrm{rep}}\mid \theta\right) \cdot p(\theta \mid y) \ \textrm{d}\theta.$

If a model fits well, test statistics $s(\cdot)$ should take similar values in both original and replicated data sets. This is summarized by a Bayesian $p$-value-like statistic, $\Pr[s(y^{\textrm{rep}})\ge s(y) \mid y] = \int \text{I}(s(y^{\textrm{rep}})\ge s(y)) \cdot p(y^{\textrm{rep}} \mid y) \ \textrm{d}y^{\textrm{rep}}.$ Any choice of statistic guided by the quantities of interest in the model itself can be used. We focus on marginal positive votes per rater and per item as statistics, because these directly reflect rater-specific biases (sensitivity and specificity) and item-level difficulty. They are simple to interpret, align with classical $\chi^2$ tests, and effectively reject models that omit difficulty when items show variation in difficulty. 

\paragraph{Leave-one-out cross-validation} We use an accurate approximation of leave-one-out cross-validation (LOO) for predictive accuracy \citep{vehtari2017practical}. LOO provides a fine-grained view of predictive performance, especially useful for model comparison and refinement.

LOO estimates out-of-sample predictive fit by evaluating the model's performance on each data point, leaving out one observation at a time. This provides an accurate measure of how well the model generalizes to unseen data. The expected log pointwise predictive density (elpd\textsubscript{loo}) is computed as $\text{elpd}_{\text{loo}} = \sum_{i=1}^{n} \log p(y_i \mid y_{-i}),$ where $p(y_i \mid y_{-i})$ is the predictive density of observation $y_i$, excluding the $i$-th observation. This involves estimating $p(y_i \mid y_{-i}) = \int p(y_i \mid \theta) \, p(\theta \mid y_{-i}) d\theta,$ which can be calculated very efficiently using Pareto-smoothed importance sampling  given a single model fit \citep{vehtari2017practical}.

\section{Extensions}\label{apd:extension}

\paragraph{Including covariate information}
Parameterizing on the log odds scale makes it straightforward to add features (i.e., covariates) in addition to the baseline random effects for items or raters.  The simple models presented here can be thought of as intercept-only versions of more general models. 

Features for raters might include demographic information such as education, location, native language, age, education level, whether they are an AI and which one, etc.  Features for items can be used to inform difficulty, discrimination and guessability.  For example, a covariate might indicate the number of options in a multiple choice test, the length of sentences used for inference, or the grade level of the textbook from which the item was culled, when a person's last dental checkup was, etc.

If item-level or rater-level covariates are available, they may be
used to inform the parameters in the usual way through a regression in the form of a
generalized linear model \citep{gelman2007data}.  For example, suppose there are item-level covariates $x_i \in \mathbb{R}^K$.  With a parameter vector $\gamma \in \mathbb{R}^K$, the generative model for the category of an item may be extended to a logistic regression,
$z_i \sim \textrm{bernoulli}\!\left( x_i^\top \cdot \gamma \right).$

\paragraph{$K$-way categorical rating} A natural extension is to $K$-way categorical ratings, such as classifying a dog image by species, classifying an article in a newspaper by topic, rating a movie on a one to five scale, classifying a doctor's visit with an ICD-10 code, and so on.  Most of the work on ratings has been in this more general categorical setting.  With more than two categories, sensitivity and specificity are replaced with categorical responses based on the latent true category.  Discrimination and guessing act the same way, but difficulty must be replaced with a more general notion of a categorical item level effect, which may represent either focused alternatives (e.g., a border collie is confusable with an Irish shepherd) or diffuse (e.g., can't tell what's in the image).

\paragraph{Population-level models} With enough raters, these models may also be extended hierarchically to make population-level inferences about the distribution of rater abilities or item difficulties \citep{paun2018}.  Several of the crowdsourcing tasks may be combined to select raters and items to rate online with active learning, which is a form of reinforcement learning. With a hierarchical model, inference may be expanded to new raters \citep{paun2018}.

\paragraph{Ordered, count, and other data} It is also straightforward to extend a rating model to ordered responses such as Likert scales \citep{teh2013infer,rogers2010semi,shatkay2005searching}, rank ordering \citep{chen2013pairwise,rafailov2024}, counts (the ``textbook'' case of crowdsourcing is estimating the number of jelly beans in a container \citep{surowiecki2005wisdom}), proportions/probabilities, distances, or pairs of real numbers such as planetary locations \citep{smyth1994}.  All that needs to change is the response model and the representation of the latent truth---the idea of getting noisy ratings and inferring a ground truth remains.  As an example, Smyth had raters mark images of Venus for volcano locations \citep{smyth1994}.  The true location is represented as a latitude and longitude and rater responses can be multivariate normal centered around the true, but unknown, location.   For ordinal ratings, an ordinal logit model of the truth may be used \citep{rogers2010semi}.  For comparisons, the Bradley-Terry model can be used \citep{bradley1952rank}, and for ranking, the Plackett-Luce generalization \citep{plackett1975analysis,luce1959individual}, as used in direct preference optimization for fine-tuning large language models \citep{rafailov2024}.

\paragraph{Joint estimation of a classifier} When item-level covariates are available, \cite{raykar2010} provide an approach to jointly estimating the parameters of the rating model and the classifier.  In essence, the prevalence model is updated to a logistic regression classifier and the the classifier participates jointly in rating along with the raters.

\newpage
\section{Full evaluation results}\label{apd:fifth}
\begin{figure}[H]
  \centering
  \includegraphics[width=1.0\linewidth]{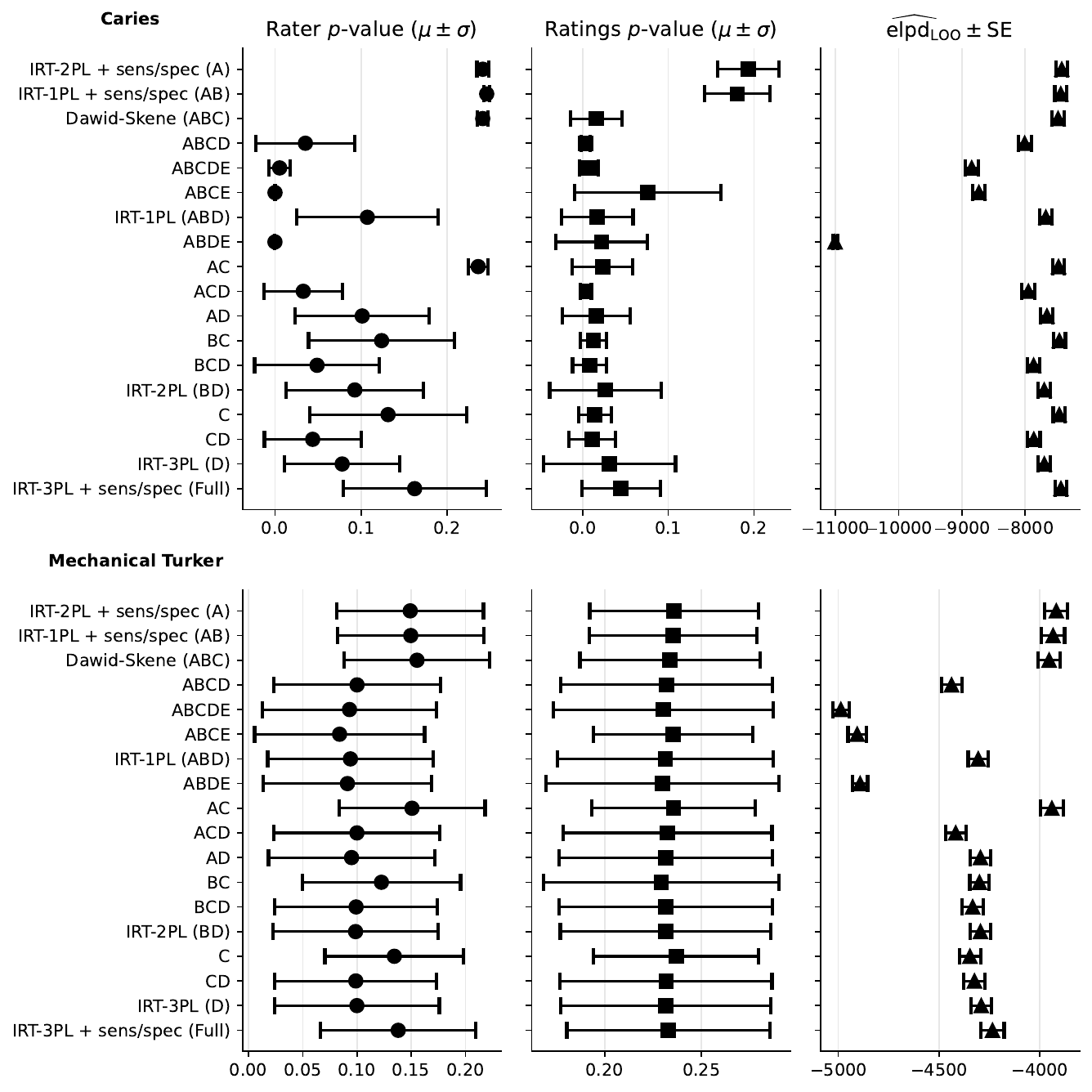}
  \caption{Posterior predictive $p$-values and leave-one-out cross-validation $\text{elpd}_{\text{loo}}$ across Caries and Mechanical Turk datasets for different 18 models. Posterior predictive values are summarized by mean and standard deviation, while $\text{elpd}_{\text{loo}}$ values are reported with standard errors.}
  \label{fig:ppc_loo_2}
\end{figure}

\end{document}